\documentclass[letterpaper]{article} % DO NOT CHANGE THIS
\usepackage[]{aaai23}  % DO NOT CHANGE THIS
\usepackage{times}  % DO NOT CHANGE THIS
\usepackage{helvet}  % DO NOT CHANGE THIS
\usepackage{courier}  % DO NOT CHANGE THIS
\usepackage[hyphens]{url}  % DO NOT CHANGE THIS
\usepackage{graphicx} % DO NOT CHANGE THIS
\usepackage{natbib}  % DO NOT CHANGE THIS AND DO NOT ADD ANY OPTIONS TO IT
\usepackage{caption} % DO NOT CHANGE THIS AND DO NOT ADD ANY OPTIONS TO IT
\usepackage{xcolor}
\usepackage{amsfonts}
\usepackage{multirow}
\usepackage{booktabs}
\usepackage{enumitem}
\usepackage{amsmath}
\usepackage{balance}
\usepackage{subcaption}

\usepackage{algorithm}
\usepackage{algorithmic}

\usepackage{newfloat}
\usepackage{listings}
\DeclareCaptionStyle{ruled}{labelfont=normalfont,labelsep=colon,strut=off} % DO NOT CHANGE THIS

\title{RIPPLE: Concept-Based Interpretation for Raw Time Series Models in Education}

\author {
    Mohammad Asadi\textsuperscript{\rm 1},
    Vinitra Swamy\textsuperscript{\rm 1},
    Jibril Frej\textsuperscript{\rm 1},
    Julien Vignoud\textsuperscript{\rm 1},
    Mirko Marras\textsuperscript{\rm 2},
    Tanja Käser\textsuperscript{\rm 1}
}
\affiliations {
    \textsuperscript{\rm 1} EPFL\\
    \textsuperscript{\rm 2} University of Cagliari\\
    mohammad.asadi@epfl.ch, vinitra.swamy@epfl.ch, jibril.frej@epfl.ch, \\ julien.vignoud@epfl.ch, mirko.marras@acm.org,
    tanja.kaser@epfl.ch
}

\usepackage{bibentry}
\begin{document}

\maketitle

\begin{abstract}

Time series is the most prevalent form of input data for educational prediction tasks. The vast majority of research using time series data focuses on hand-crafted features, designed by experts for predictive performance and interpretability. However, extracting these features is labor-intensive for humans and computers. In this paper, we propose an approach that utilizes irregular multivariate time series modeling with graph neural networks to achieve comparable or better accuracy with raw time series clickstreams in comparison to hand-crafted features. Furthermore, we extend concept activation vectors for interpretability in raw time series models. We analyze these advances in the education domain, addressing the task of early student performance prediction for downstream targeted interventions and instructional support. Our experimental analysis on 23 MOOCs with millions of combined interactions over six behavioral dimensions show that models designed with our approach can (i) beat state-of-the-art educational time series baselines with no feature extraction and (ii) provide interpretable insights for personalized interventions.
Source code: \url{https://github.com/epfl-ml4ed/ripple/}.

\end{abstract}

\section{Introduction}

% \begin{itemize}
%     \item What is the problem and why is it important?
%     \item What have others done to solve that problem?
%     \item What is the gap/what is missing?
%     \item What is the contribution of this paper (how are we contributing to this gap)?
% \end{itemize}
Over the last three years, there has been a 10-fold increase in digital learners on massive open online courses (MOOCs), contributing to a popular and data-rich setting in education \cite{impey2021moocs, ByTheNum39}. In enabling a completely online learning experience, MOOCs suffer from high dropout and low success rates \cite{aldowah2020factors}. Thus, an important task to counter these phenomena is providing personalized guidance at scale \cite{perez2021can}. This task requires (i) predicting student performance early enough to intervene and adjust learning pathways and (ii) interpreting which behavior contributes to failing and passing trajectories for each student. 

There exists a large body of approaches on student success prediction in MOOCs, e.g., random forests \cite{marras2021can, sweeney2016next}, logistic regression \cite{whitehill2017mooc}, or neural networks \cite{wang2017deep, mu21deeplearning}. Most of these methods operate post-hoc, i.e., in the context of the entire time series. Only few works have focused on predicting success early on during the course. For example, \citet{mbo20early} predicted students' pass-fail grades based on video interactions, while \citet{mao2019one} used temporal patterns to intervene early in programming tasks. Most of the work has employed hand-crafted expert-designed features, ranging from engagement-based features, such as course attendance rates \cite{he18utilize} or the number of online sessions \cite{chen2020utilizing, lemay2020grade}, to features capturing fine-grained video behavior \cite{DBLP:conf/edm/AkpinarRA20, mu21deeplearning} and measuring students' learning regularity \cite{boroujeni2016quantify}. Recently, \citet{marras2021can} performed a meta-analysis on early success prediction features, showing that their predictive power does not often generalize across courses and raising questions about which features should be selected based on the course characteristics. Designing and extracting features for educational time series hence becomes expensive in terms of human and computational resources. Minimal literature has addressed raw time series in education. \citet{prenkaj2021hidden} used auto-encoders for risk prediction in MOOCs, but did not provide comparisons to hand-crafted baselines. 

%Notably, when raw time series is used in education, there are no instances of student-level model interpretability analysis. More generally, using raw time series data instead of hand-crafted features leads to less interpretable models.
%A few papers have attempted to use the latent dimensions constructed by auto-encoders to avoid the human labor cost of hand-crafted features \cite{autoencoders19}; however, our preliminary experiments of reconstructing sparse, irregular, and multivariate time series with auto-encoders were not able to converge. In this paper, we use the state-of-the-art methods from recent AI conferences in transformers \cite{transformer}, set functions \cite{seft}, and sensor-based graph neural networks \cite{zhang2021graph} to perform early success prediction using raw time series. 

Using raw time series in combination with neural networks has also led to black-box models. In response to this issue, there has been a strong increase in research on neural network explainability, with methods such as LIME \cite{lime}, SHAP \cite{shap}, and counterfactual explanations \cite{cem}. Only few works have however focused on explainability in the domain of education. Prior research has used LIME to provide local explanations for performance prediction models \cite{hasib2022lime, vultureanu2021improving} or to build a basis for students dashboards \cite{scheers2021interactive}. \citet{baranyi2020interpretable} applied SHAP to interpret student dropout prediction models. However, a major shortcoming of those methods is that they do not seem to agree about what features are important in MOOCs \cite{swamy2022explainability}. Furthermore, interpretations are limited only to the engineered features originally during model training. On raw time series predictions, the minimal existing literature has shown attention heatmaps for temporal insights, not higher level, human-friendly actionable features for educator interventions \cite{ts_interp}.

In this paper, we propose \texttt{Ripple} (Raindrop InterPretability PipeLine for Education), a novel methodology for providing interpretable early student success prediction using raw time series data. In contrast to prior work, our pipeline does not require any feature engineering, while still providing accurate predictions as well as human-friendly explanations. Our pipeline is based on the combination of a graph-based neural network approach \cite{zhang2021graph} for classifying raw time series of student interactions and the adaptation of concept activation vectors (TCAV) \cite{kim2018interpretability} for interpreting the neural network's internal state. Specifically, we use six well-defined dimensions of self-regulated learning in online courses from recent literature \cite{mejia2022identifying} to provide interpretability in the global and local context. To the best of our knowledge, TCAV has never been applied on time series.
We evaluate our pipeline on a large educational data set including $23$ MOOCs with over $100,000$ students and millions of interactions, addressing the following research questions: \begin{enumerate}
\small
    \item Can we use raw time series as input and achieve comparable performance to hand-crafted features?
    \item Can we obtain interpretability on raw multivariate time series through learner-centric concept activation vectors?
\end{enumerate}
Our results show that graph neural networks allow us to achieve comparable or better performance with raw time series models to hand-crafted features in $18$ out of $23$ courses and beat other state-of-the-art time series baselines on $21$ out of $23$ courses. Moreover, we showcase our interpretable pipeline on a selected digital signal processing course.

\begin{figure*}[!t]
  \centering
  \includegraphics[width=0.95\textwidth, trim=1 1 1 1,clip]{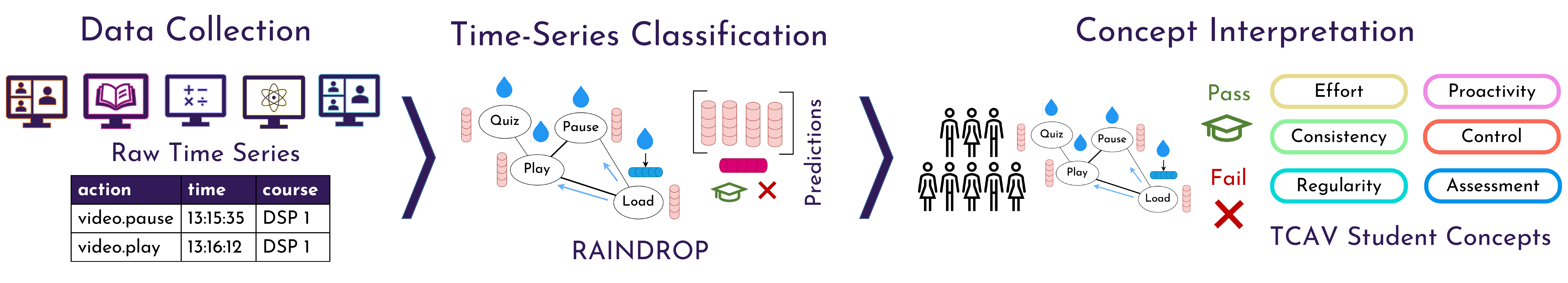}
  \caption{Our \texttt{Ripple} time series interpretability approach from logs collection to concept vector analysis.}
  \vspace{-3mm}
  \label{fig:features}
\end{figure*}

% %(related work embedded)
% We aim to answer the following research questions: \\
% 1. Can we use raw time series as input and achieve comparable performance to hand-crafted features? \\
% 2. Can we obtain interpretability on raw multivariate time series through learner-centric concept activation vectors? \\
% 3. Can we use interpretability as a basis for building trust in model performance?

% With our pipeline we...
% The motivation for naming the pipeline \texttt{Ripple} comes from the underlying \texttt{Raindrop} time series model, which is named for the effect of a single interaction (a raindrop) dispersing through a graph (a body of water). To interpret the effects of this drop is to describe the \texttt{Ripple} of the interaction.

\section{Methodology}

This paper targets a classification task that utilizes raw multivariate time series to predict student pass-fail labels \textit{early} in a course. Our goal is to achieve at least comparable performance using raw time series data in comparison to hand-crafted features (e.g., \citet{marras2021can}), without compromising on interpretability. We first formalize the posed problem and then describe our methodology.

\subsection{Problem Formalization}

% \jibril{temporary list of notations (may need to add more depending on the \texttt{Raindrop} and TCAV sections) + change concept notation?, will be changed into a paragraph soon}

% \begin{itemize}
%     \item $z_s^c$ \texttt{Raindrop} embedding computed from $I_s^c$
%     \item $Xo$ concept
%     \item $P_{Xo}$ positive set of examples for explainable concept $Xo$
%     \item $N_{Xo}$ negative set of examples for explainable concept $Xo$
%     \item $v_{Xo}$ Concept Activation Vector (CAV) for concept $Xo$
% \end{itemize}

Given a course $c$ part of the offering $\mathbb{C}$, we denote as $\mathbb S^c$ the set of students enrolled in $c$. Since each course can be run multiple times, we define a course set $\tilde{\mathbb C} = \{c_1, \ldots, c_{M^{\Tilde{\mathbb C}}}\} \subset \mathbb C$ as the set of all iterations of the same course over the years, with $M^{\tilde{\mathbb C}}$ being the total number of iterations for the course set $\tilde{\mathbb C}$. Each course $c$ includes a set of $N_c$ learning objects denoted as $\mathbb{O}^c$. Students interact with learning objects in $\mathbb{O}^c$. The interactions of a student $s \in \mathbb S^c$ are modeled as a time series $\mathbb I_s^c = \{i_1, i_2, \ldots\}$. Each interaction is represented with a tuple composed of a timestamp $t$, an action $a$, a learning object $o \in \mathbb{O}^c$, and optional metadata $m$, i.e., $i = \left(t, a, o, m \right)$. We denote as $y_s^c \in \{0,1\}$ the pass-fail label for student $s$ in course $c$. Training a classification model $\mathcal M_\theta : \mathbb I \xrightarrow{} \{0, 1\}$ is an optimization problem aimed to minimize the expectation on the following objective function:
%expectation on the following objective function:

% \begin{equation}
% \label{eq:problem-definition}
% \resizebox{0.9\hsize}{!}{%
% $\tilde{\mathcal M_\theta} = \underset{\mathcal M_\theta}{\operatorname{argmin}} \mathop{\mathbb{E}}_{s \; \in \; \mathbb S^c} \mathcal M_\theta(\mathbb I_s^c) - y_s^c$}
% \end{equation}

\begin{equation}
\tilde{\mathcal M_\theta} = \underset{\mathcal M_\theta}{\operatorname{argmin}} \mathop{\mathbb{E}}_{s \; \in \; \mathbb S^c} | \mathcal M_\theta(\mathbb I_s^c) - y_s^c|
\label{eq:problem-definition}
\end{equation}

To preserve transparency, we assume that the prediction $\tilde{y}_s^c = \tilde{\mathcal M_\theta}(\mathbb I_s^c)$ for student $s$ in course $c$ can be interpreted based on a set $\mathbb P$ of human-understandable educational concepts. Each concept $p \in \mathbb P$ is associated to a relative concept importance score $d^{c,p,y}$ ranging in $[0, 1]$. A value close to $0$ ($1$) means that the concept $p$ has a low (high) importance for the $y$-class model predictions based on the interactions $\mathbb I^c$. An example concept is student's \textit{regularity} in the course.
     
Following this formalization, we devised our deep learning approach consisting of the three main stages illustrated in Fig. \ref{fig:features}: (i) data collection and preprocessing ($\mathbb I_s^c$), (ii) raw time series classification ($\tilde{\mathcal M_\theta}$), and (iii) concept-based interpretation ($\mathbb P$). We discuss each stage in more detail.

% and as $\hat{y}_s^c \in \{0,1\}$ the pass-fail label prediction.  
    
% \jibril{Is it necessary in this paper to separate action and learning object? It would make thinks easier for \texttt{Raindrop} description if we consider an interaction as a tuple (t, a) with a an action} \vini{I think we need the learning object because we refer to Problem.ID and Video.ID in the \texttt{Raindrop} input.}

\subsection{Raw Time Series Collection and Preprocessing}\label{sec:preproc} 
% What does \texttt{Raindrop} input require
\vspace{1mm} \noindent \textit{\textbf{Collection}}.  We collected clickstream data involving interactions $\mathbb I$ for students $\mathbb S^c$ from MOOCs $c \in \mathbb C$, modelled as an irregular multivariate time series. We refer to our time series as irregular due to the non-uniform time interval the data was generated (e.g., a student did not interact for over a week). Multivariate refers to the learning objects involved in the actions $a \in \mathbb A$, used to model the time series $\mathbb I$. 

To model each interaction $i = \left(t, a, o, m \right)$, we considered learning objects $\mathbb{O}^c$ of type video and problems, with the video actions $\mathbb A_v \subset \mathbb A$ = \{Download, Error, Load, Pause, Play, Seek, SpeedChange, Stalled\} and the problem actions $\mathbb A_p \subset \mathbb A$ = \{IsAssignment, IsQuiz\}. An ID was assigned to each video and problem. For each problem, the number of times it was attempted by the student (Problem SubmissionNum) was also tracked. Each timestamp $t \in \mathbb N$ and action $a \in \mathbb A$, alongside the metadata $m$ of Video ID, Problem ID, and Problem SubmissionNum, were treated as separate variables in our time series. For brevity, we will refer to our irregular multivariate time series as simply raw time series.

\vspace{1mm} \noindent \textit{\textbf{Early-Dropout Filtering}}. A common archetype of MOOC student is a learner who watches only a few videos or makes only a few initial interactions ($\mathbb I_s^c$). Motivations for this behavior include misaligned expectations of course material, unexpected life circumstances, or intellectual curiosity for a small subset of videos \cite{onah2014dropout, goopio2021mooc}. These students can be easily predicted with fail labels ($y_s^c = 1$) by considering their (lack of) initial graded assignments using a simple logistic regression model. It does not make sense to pass this student subset to complex neural networks when they could be so concisely and accurately identified as failing without further analysis. To let our raw time series model focus on hard-to-identify students, we removed students which can be predicted as failing with $99\%$ accuracy using two weeks of assignment data, via the same model proposed in \cite{swamy2022meta, swamy2022explainability}\footnote{These models were optimized through a grid-search to determine the optimal accuracy threshold and number of weeks.}. In the rest of this paper, we will refer to $\mathbb S^c$ as the set of students after this filtering.

\vspace{1mm} \noindent \textit{\textbf{Early Prediction Level Definition}}. To enable our model to support instructors during the course \cite{borrella2021taking, xing2019dropout, whitehill2015beyond}, we considered an \textit{early} prediction setting. Under an early prediction level $e \in [0, 100]$ representing the percentage of the course duration at which the prediction is delivered, we considered interaction data only up to that point in time. For instance, if $e = 60\%$ and the course lasts $10$ weeks, we would consider only interactions happening in the first six course weeks. We denote the interactions of $s$ in $c$ up to the early prediction level $e$ as $\mathbb I_s^{c,e}$. 

\subsection{Raw Time Series Classification}

\begin{figure}[!t]
    \centering
    \includegraphics[width=0.47\textwidth]{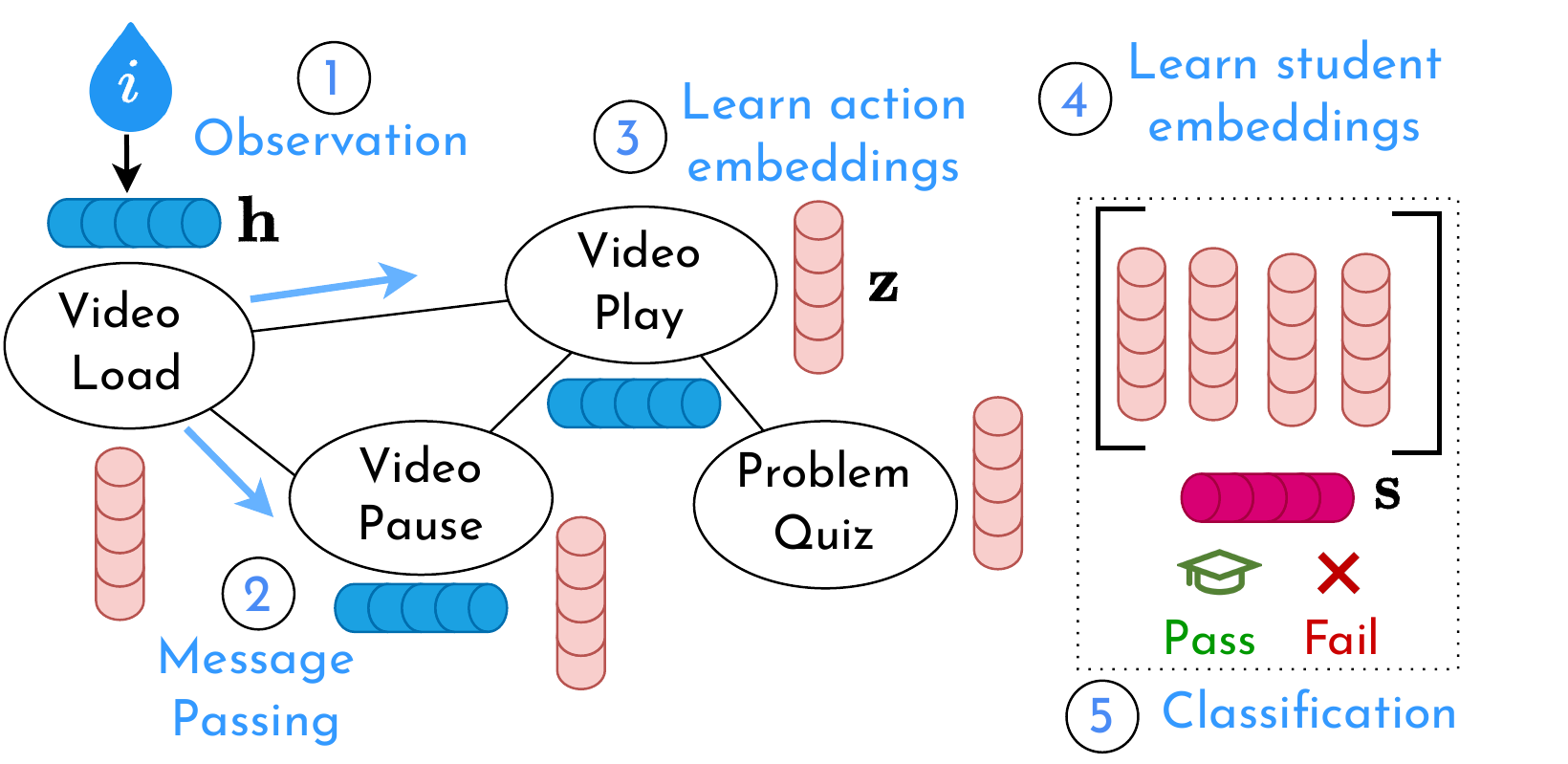}
    \caption{\texttt{Raindrop} time series classification approach for educational data: (1) observe an action and obtain interaction embeddings $\textbf{h}$; (2) use message passing to compute interaction embeddings for the unobserved actions; (3) learn action embeddings $\textbf{z}$ from the interaction embeddings; (4) learn student embeddings $\textbf{s}$ from the action embeddings; (5) course failure classification from student embeddings.}
    \vspace{-3mm}
    \label{fig:raindrop}
\end{figure}

%\jibril{need to change the vocabulary used, e.g. action, interaction, variable still mixed buy same meaning here}

\vspace{1mm} \noindent \textit{\textbf{Motivation}}. The irregularity and multivariate nature in our time series is generally hard to analyze using classical machine learning temporal models that assume fully (or regularly) observed fixed-size inputs \cite{ismail2019deep}. To counter these issues, a recent time series representation model (\texttt{Raindrop}) assumes that actions are dependent and leverages their hidden structure by using a directed weighted dependency graph~\cite{zhang2021graph}. When an action $a \in \mathbb A$ is observed within an interaction $i \in \mathbb I$, this model updates $a$'s internal representation and uses the dependency graph to update those of the actions related to $a$. The intuition is that an action observed at timestamp $t$ can imply how unobserved actions would behave; updating these unobserved actions can improve the time series modeling.

\vspace{1mm} \noindent \textit{\textbf{Task Definition}}. Given an irregularly sampled multivariate time series $\mathbb I^c$, where each sample $\mathbb I^c_s$ has multiple but not always observed actions and each action has a different number of observations, our model $\tilde{\mathcal M_\theta}$ first learns a function $\mathcal F : \mathbb I^c_s \xrightarrow{} \mathbb R^{*}$ that maps $\mathbb I^c_s$ to a fixed-length representation $\textbf{s}_s^c$ (student embedding) suitable for classification. Using learned $\textbf{s}_s^c$, this model then predicts the label $\tilde{y}_s^c$. The learned representation captures temporal patterns of irregular observations and considers dependencies between actions. 

\vspace{1mm} \noindent \textit{\textbf{Model Learning}}. To learn the representation $\textbf{s}_s^c$ for student $s$ in a course $c$, we implemented the \texttt{Raindrop} architecture described in \citet{zhang2021graph}, generating student-level embeddings using a hierarchical architecture composed of three levels aimed to model interactions, actions, and students (see Fig. \ref{fig:raindrop}). First, we built a dependency graph $\mathcal{G}_s$ for every student $s$, where nodes represent actions and directed edges indicate the relation (with a weight ranging in $[0, 1]$) between two actions. The edge weights were initialized to $1$ and optimized student-wise and time-wise via message passing, starting from the node associated to the observed action. 

When an interaction $i = \left(t, a, o, m \right)$ was fed into the model for student $s$ at time $t$, the model first embedded the interaction for the observed action (i.e., the action whose value was recorded) in an interaction embedding $\textbf{h}$ using a non-linear transformation of the input. In order to update the interaction embeddings for unobserved actions at timestamp $t$, a graph neural network was used on top of the dependency graph $\mathcal{G}_s$. Once the interactions embeddings were generated from all timestamps, temporal self attention was used to aggregate all interactions embeddings associated to a given action into a single fixed-size representation $\textbf{z}$. The student embedding $\textbf{s}_s^c$ was obtained by concatenating all action embeddings $\textbf{z}$. The final classifier is a fully-connected network that received $\textbf{s}_s^c$ and output the pass-fail label $\tilde{y}_s^c$.

\subsection{Concept-Based Model Interpretation}
% TCAV methodology
% How we modified TCAV to work with Time Series

% Interpretation is important in education
\vspace{1mm} \noindent \textit{\textbf{Motivation}}. Recent deep learning models, like \texttt{Raindrop}, trade transparency for accuracy. 
However, social, ethical and legislative requirements prominently call for model transparency, especially in education \cite{webb2021machine,conati2018ai}. 
Identifying the possible reasons behind a predicted failure in addition to predicting it accurately is crucial for designing effective interventions.
A popular approach to interpretability is the use of post-hoc explainability methods, which return importance scores in terms of the input features the model originally considered. However, these methods appear ineffective on models receiving raw time series. %In our setting, the input raw time series does not respond to high-level concepts that humans easily understand and the model’s neural activations seem incomprehensible. 
Because of this difficulty, there is a need to shift towards learner-centric concept explanations.  
With this in mind, we adopt \citet{kim2018interpretability}'s human-friendly quantitative testing based on concept activation vectors (TCAV), which gives an interpretation of a neural network’s internal state in terms of human-interpretable concepts not explicitly considered as an input feature by the model.
Used primarily for image and occasionally text data, this technique has never been used on (educational) time series input, to the best of our knowledge. 
The strengths of TCAV lay in its flexibility to analyze whichever concepts an educator finds pertinent for their course setting. %For example, an educator might want to specify a concept of high performance on a certain unit of the course (e.g., weeks three through five) or high engagement on forums.

\vspace{1mm} \noindent \textit{\textbf{Concept Design and Extraction}}. To extract concepts for our educational scenario, we used the six learning dimensions (see Table \ref{tab:profiles}) proposed by \citet{mejia2022identifying} due to the similarity of the underlying course data, the ease in interpreting these profiles, the clear identification of actionable insights based on patterns in these dimensions, the underpinning educational theory validated in them, and their relationship with academic performance. Concerning \textit{effort}, \textit{control}, and \textit{assessment}, \citet{mejia2022identifying} found differences among profiles in terms of intensity (higher / lower). The \textit{consistency} dimension was found to capture differences in the relative intensity over the course, with the majority of students having small peaks (uniform) and only a few students working more in the first/last course weeks (first / last half). Regarding \textit{regularity}, some students were found to regularly work on specific weekdays (higher peaks), while others did not have a clear pattern (lower peaks). For \textit{proactivity}, most students were found to interact with the content in advance (anticipated), whereas a minority often interacted with it after the deadline (delayed). 

\begin{table}[!t]
\centering
\small
\setlength{\tabcolsep}{3pt}
\resizebox{\hsize}{!}{%
\begin{tabular}{rll}
\toprule
\textbf{Dimensions}   & \multicolumn{1}{l}{\textbf{Measures}} & \multicolumn{1}{l}{\textbf{Patterns}}                                                                                                   \\ \midrule
\textbf{Effort}      & \begin{tabular}[c]{@{}l@{}}  Total time online\\ Total video clicks\end{tabular}  &  \begin{tabular}[c]{@{}l@{}} Higher intensity \\ Lower intensity \end{tabular}                                        \\
\midrule
\textbf{Consistency} & \begin{tabular}[c]{@{}l@{}} Mean session duration\\ Relative time online\\ Relative video clicks\end{tabular}    &    \begin{tabular}[c]{@{}l@{}} Uniform \\ First half \\ Second half \end{tabular}     \\
\midrule
\textbf{Regularity}  & \begin{tabular}[c]{@{}l@{}} Periodicity of week day\\ Periodicity of week hour\\  Periodicity of day hour\end{tabular}  & \begin{tabular}[c]{@{}l@{}} Higher peaks \\ Lower peaks \end{tabular}  \\
\midrule
\textbf{Proactivity} & \begin{tabular}[c]{@{}l@{}} Content anticipation\\  Delay in lecture view\end{tabular}                                       &  \begin{tabular}[c]{@{}l@{}} Anticipated \\ Delayed \end{tabular}  \\
\midrule
\textbf{Control}     & \begin{tabular}[c]{@{}l@{}} Fract. time spent (video)\\  Pause action frequency\\  Average change rate\end{tabular} & \begin{tabular}[c]{@{}l@{}} Higher intensity \\ Lower intensity \end{tabular}  \\
\midrule
\textbf{Assessment}  & \begin{tabular}[c]{@{}l@{}} Competency strength\\ Student shape\end{tabular}  & \begin{tabular}[c]{@{}l@{}} Higher intensity \\ Lower intensity \end{tabular}  \\ \bottomrule
\end{tabular}
}
\caption{Learning dimensions from \citet{mejia2022identifying} used as concepts for interpretability in our study.}
\vspace{-3mm}
\label{tab:profiles}
\end{table}

We derived our set of concepts $\mathbb P$ from \citet{mejia2022identifying}'s findings, using the above patterns emerged per dimension. For each dimension, we identified two or three student patterns (e.g., the subset of students showing the highest \textit{effort} and the subset of students showing the lowest \textit{effort}) and devised a greedy optimization protocol to select approximately 100 students that most fit the considered pattern. We achieve this by extracting the $t = 5\%$ of top students showing the considered student pattern for each \textit{corresponding measure} in that dimension (see Table \ref{tab:profiles}) and computing the intersection of these measure subsets. We then incrementally increased the threshold $t$ until each combined pattern subset had at least 100 students\footnote{We have experimentally validated that TCAV is not consistent or robust with less than 100 examples.}. 

%%%%%%%%%%Table listing all the courses%%%%%%%%%%%%%%%%%%%%%%%%%%%%%%%%%%
\begin{table*}[!t]
\small
\resizebox{\textwidth}{!}{
\begin{tabular}{lllrrllrrr}
\toprule
\textbf{Course Title} & \textbf{Identifier} & \textbf{Field$^1$} &  \textbf{It.} & \multicolumn{1}{r}{\textbf{\begin{tabular}[c]{@{}c@{}}No. \\ Stud.$^2$\end{tabular}}} & \textbf{Level} & \textbf{Lang.} & \multicolumn{1}{r}{\textbf{\begin{tabular}[c]{@{}c@{}}No. \\ Weeks\end{tabular}}} &  \multicolumn{1}{r}{\textbf{\begin{tabular}[c]{@{}c@{}}Passing \\ Rate$^3$\end{tabular}}} & \multicolumn{1}{r}{\textbf{\begin{tabular}[c]{@{}c@{}}No. \\ Quiz.$^4$\end{tabular}}} \\
\midrule
CPP Programming & \textit{CPP} & CS & 2 & 1,517 & Prop. & En/Fr & 8/10 & (38, 63) & 12 \\
Digital Signal Processing & \textit{DSP} & CS & 5 & 15,394 & MSc & English & 10 & (17, 24) & 38 \\
Functional Programming & \textit{ProgFun} & CS & 2 & 18,702 & BSc & French & 7 & (52, 82) & 3 \\
Analyse Numérique & \textit{AnNum} & Math & 3 & 1,468 & BSc & French & 9 & (9, 75) & 36 \\
Éléments de Géomatique & \textit{Geomatique} & Math & 1 & 452 & BSc & French & 11 & 45 & 27\\
Household Water Treatment & \textit{HWTS} & NS & 2 & 2,423 & BSc & French & 5 & (46, 49) & 10 \\
Microcontrôleurs & \textit{Micro} & Eng & 4 & 7,503  & BSc & French & 10 & (8, 49) & 18 \\ 
Launching New Ventures & \textit{Venture}  & Bus & 1 & 3,208  & BSc & English & 7 & 3 & 13\\
Villes Africaines & \textit{VA} & SS & 3 & 10,094 & BSc/Prop. & En/Fr & 12  & (8, 11) & 18 \\
\bottomrule
\end{tabular}}
\normalsize $^1$\textbf{Field.} \textit{Bus}: Business; \textit{CS}: Computer Science; \textit{Eng}: Engineering; \textit{Math}: Mathematics; \textit{NS}: Natural Science; \textit{SS}: Social Science. \\
\normalsize $^2$\textbf{No. Students} is calculated after filtering out the early-dropout students, as detailed in the \textit{Time Series Preprocessing} section. \\
\normalsize $^3$\textbf{Passing Rate} is the (min, max) of passing rate percentage over iterations. $^4$\textbf{No. Quizzes} is the average number of quizzes. 
\caption{Detailed information on the MOOCs highlighted in our experiments.}
\vspace{-3mm}
\label{tab:courses}
\end{table*}

\vspace{1mm} \noindent \textit{\textbf{Concept Importance Computation}}. 
For a given dimension, the two identified student subsets were given as an input to TCAV, which relied on them to (i) identify a hyperplane that best differentiates between the model activations produced by the subset and the activations in any model layer, and (ii) specify a CAV, i.e., the direction orthogonal to this hyperplane. Using the CAV directional derivative, we identified the importance score of each concept for the predictions our model returned.
Formally, let $y$ and $\mathbb S_y$ represent the pass-fail label and the set of students with that label respectively, and let $\mathcal D^{p, l}$ be the directional CAV derivative function for concept $p \in \mathbb P$ at the model layer $l \in \mathbb L$. The TCAV importance score for $p$ is the fraction of $y$-class students whose activation vector was on average positively impacted by $p$:

$$
\operatorname{d}^{c, p, y}= \frac{1}{|\mathbb L|} \sum_{l \; \in \; \mathbb L} \frac{\left|\left\{s \in \mathbb S_y: \mathcal D^{p, l}(s)>0\right\}\right|}{\left|\mathbb S_y\right|} 
$$

TCAV importance scores range between $[0,1]$. Higher values indicate that concept $p$ has a high importance for the prediction of class $y$. The sensitivity of concepts to predictions can be specified for a population of students (global interpretation) or for individual students (local interpretation). 

%%%%%%%%%%%%%%%%%%%%%%%%%%%%%%%%%%%%%%%%%%
%Result Section
%%%%%%%%%%%%%%%%%%%%%%%%%%%%%%%%%%%%%%%%%%
\section{Experimental Evaluation}
We examined whether models using raw time series as input can achieve comparable performance to models receiving hand-crafted features (RQ1) and whether we can obtain interpretable concept activation vectors to gain insights into model predictions (RQ2). In the following, we describe the dataset, optimization protocol, and the experiments in detail.

%%%%%%%Description of data set%%%%%%%%%%%%%%%  
\vspace{1mm} \noindent \textit{\textbf{Dataset}}. Our dataset consisted of $23$ MOOCs and $134{,}699$ students. Its entries are fully anonymized and correspond to courses offered by an European university worldwide between 2013 and 2015. Facets of this dataset have been used in educational machine learning work, such as \citet{swamy2022meta, swamy2022explainability, mejia2022identifying, li2015video, boroujeni2018discovery}. The dataset records include fine-grained video and quiz interactions for each student, e.g., pressing pause on a video or submitting a quiz. After early-dropout filtering, our data set included $73{,}042$ students in total. The $23$ courses were selected from a set of larger MOOC courses for diversity in topic, duration, level, language, and student population, which allowed us to provide a realistic estimation of model performance. The course size ranges from $452$ to $11{,}151$ students. Table \ref{tab:courses} lists detailed course information.

%%%%%%%%%%%%%%%
%%%% Subsection OPTIMIZATION
%%%%%%%%%%%%%%%%%
\vspace{1mm} \noindent \textit{\textbf{Optimization Protocol}}. To answer our research questions, we compared the performance of our model to the optimal bidirectional LSTM (BiLSTM) architecture using hand-crafted features, both described in \citet{swamy2022meta}. To provide another point of comparison, we also implemented the Set Functions for Time Series (SeFT) \cite{seft} and Transformers \cite{transformer} baselines analyzed by \citet{zhang2021graph} as other state-of-the-art models in the medical domain. We trained each model on the $23$ course iterations listed in Table \ref{tab:courses} under two early prediction levels ($e \in \{40\%, 60\%\}$). Following prior work \cite{swamy2022meta}, our choice of these two levels is motivated by the fact that \textit{HWTS}, the shortest course, has only 5 weeks. We used a $80:10:10$ train-test-validation split, making sure to assign each student's time series uniquely in either train, test, or validation. We monitored balanced accuracy (BAC) due to the high class imbalance\footnote{We found metrics other than BAC (e.g., F1, AUC, precision, and recall) to show a biased perspective of model performance.}. For each model, the hyper-parameters were tuned via a grid search (please refer to our source code). 

%The BiLSTM's hyper-parameters were optimized through a grid search over hidden layer sizes $32$, $64$, and $128$. For the three other models, hyper-parameters were optimized through a grid search over TransformerEncoder layer sizes of $1$, and $2$ and classification layer sizes of $1$, and $2$. 

%%%%%%%%% Table with BAC results %%%%%%%%%%%%%%%%%%%%%%%%%%%%%%%%%%%%%%%%%
\begin{table*}[]
\small
\centering
\resizebox{1.\textwidth}{!}{
\begin{tabular}{@{}r|c|rr|rr|rr||c|rr|rr|rr@{}}
\toprule
\multicolumn{1}{l|}{} & \multicolumn{7}{c||}{\textbf{Early 40}\%} & \multicolumn{7}{c}{\textbf{Early 60}\%}\\

\textbf{} & \textbf{Raindrop}  & \multicolumn{2}{|c}{\textbf{SeFT}} & \multicolumn{2}{|c}{\textbf{TF}} & \multicolumn{2}{|c||}{\textbf{BiLSTM}} & \textbf{Raindrop}  & \multicolumn{2}{|c}{\textbf{SeFT}} & \multicolumn{2}{|c}{\textbf{TF}} & \multicolumn{2}{|c}{\textbf{BiLSTM}}\\

\textbf{} & BAC & BAC & R & BAC & R & BAC & R & BAC  &  BAC  & R &  BAC  & R &  BAC  & R\\ \midrule

CPP*& \textbf{0.57} & 0.46 & 2/2 & 0.54 & 2/2 & 0.56 & 2/2 & \textbf{0.55} & 0.53 & 1/2 & 0.52 & 2/2 & \textbf{0.55} & 2/2\\
DSP*& \textbf{0.81} & 0.72 & 5/5 & 0.59 & 5/5 & 0.80 & 4/5 & \textbf{0.91} & 0.82 & 5/5 & 0.62 & 5/5 & \textbf{0.91} & 4/5 \\
ProgFun* & \textbf{0.76} & 0.63 & 2/2 & 0.53 & 2/2 & 0.63 & 2/2 & \textbf{0.75} & 0.69 & 2/2 & 0.56 & 2/2 & 0.67 & 2/2 \\
AnNum  & \textbf{0.66} & 0.51 & 3/3 & 0.51 & 3/3 & 0.62 & 3/3 & 0.55 & 0.57 & 3/3 & 0.51 & 3/3 & \textbf{0.69} & 1/3\\
Geomatique* & 0.50 & 0.45 & 1/1 & \textbf{0.56} & 0/1 & 0.47 & 1/1 & \textbf{0.77} & 0.55 & 1/1 & 0.45 & 1/1 & 0.76 & 1/1\\
HWTS & 0.61 & 0.55 & 2/2 & 0.55 & 1/2 & \textbf{0.71} & 1/2 & 0.62 & 0.62 & 1/2 & 0.56 & 2/2 & \textbf{0.73} & 0/2\\
% Structures & 0.53 & 0.50 & 0.42 & 0.46 & 0.51 & 0.60 & 0.53 & 0.56 \\
Micro & 0.74 & 0.70 & 2/4 & 0.58 & 4/4 & \textbf{0.81} & 1/4 & \textbf{0.78} & 0.76 & 2/4 & 0.63 & 2/4 & \textbf{0.78} & 2/4\\
Ventures* & \textbf{0.77} & 0.64 & 1/1 & 0.64 & 1/1 & 0.50 & 1/1 & \textbf{0.88} & 0.73 & 1/1 & 0.56 & 1/1 & 0.60 & 1/1\\
VA* & \textbf{0.88} & 0.75 & 3/3 & 0.63 & 3/3 & 0.80 & 3/3 & \textbf{0.90} & 0.72 & 3/3 & 0.68 & 3/3 & 0.83 & 3/3\\

\bottomrule
\end{tabular}}
\vspace{0.25mm}
\\\normalsize The best model for each course type and early prediction level is marked in \textbf{bold}. Course types where \texttt{Raindrop} had comparable or better performance to BiLSTM on both early prediction levels are marked in (*).\\
\caption{Performance comparison between three raw time series models (Raindrop, SeFT, Transformers) and a  hand-crafted feature-based model (BiLSTM). Balanced Accuracy (BAC) is averaged over iterations of the same course and weighted by the number of students. R indicates the proportion of iterations of a course where \texttt{Raindrop} outperforms the baseline.}
\vspace{-3mm}
\label{tab:rq1_baclist}
\end{table*}

%%%%%%%%%%%%%%%%%%%%%%%%%%%%%%%%%%%%%%%%%%%%%%%%%%%
%%%% RESULTS RQ1
%%%%%%%%%%%%%%%%%%%%%%%%%%%%%%%%%%%%%%%%%%%%%%%%%%%
\subsection{RQ1: Raw Time Series Classification}
In a first analysis, we compared \texttt{Raindrop}'s performance to (i) state-of-the art models (Transformers, SeFT) using raw time series and (ii) a BiLSTM using $42$ features engineered for educational data. Table \ref{tab:rq1_baclist} lists the BAC for all the models and course types for both the $40\%$ and $60\%$ early prediction levels. For each course type, the BAC is averaged over the number of courses, weighted by the number of students. In our preliminary experiments, we found that LSTMs and autoencoders could not converge on raw time series input, always creating models with 50\% balanced accuracy. %The best performing model per course type is denoted in bold and an asterisk (*) marks the course types where \texttt{Raindrop} achieved at least comparable performance to BiLSTM using engineered features.

For the $40\%$ early prediction level, \texttt{Raindrop} achieves a comparable or better BAC than the state-of-the-art model with engineered features for $7$ out of $9$ course types. Moreover, \texttt{Raindrop} is overall the best model for six course types. At the level of a single course, \texttt{Raindrop} performs equally or better than Transformers for $21$ out of $23$ courses and than SeFT for $21$ out of $23$ courses. Comparing to engineered features, \texttt{Raindrop} using raw time series exhibits a higher or comparable BAC for $18$ out of $23$ courses (higher BAC: $12$ courses, comparable BAC: $6$ courses). Note that comparable is defined as a less than $5\%$ decrease in BAC. Given the effort required for engineering the features, we deem a small decrease in predictive performance acceptable.

% \tanja{Some more details/examples needed, where \texttt{Raindrop} has issues. It is mainly Microcontroleurs (2 courses), HWTS (both courses), and 1 Structures course, where \texttt{Raindrop} looses a lot compared to the other models.}
We observe similar results for the $60\%$ early prediction level. Again, using raw time series with  \texttt{Raindrop} leads to a BAC comparable (or higher) than using engineered features for $7$ out of $9$ course types. Furthermore, \texttt{Raindrop} is the best model for seven course types. At the level of single courses, \texttt{Raindrop} has equal or better performance than Transformers and SeFT for $21$ and $20$ courses, respectively. When considering both $40\%$ and $60\%$ early prediction, using \texttt{Raindrop} leads to accuracy levels higher than the BiLSTM using engineered features on $6$ out of the $9$ course types. HWTS and Micro are the only two course types where BiLSTM with hand-crafted features outperforms models that use the raw time series.

\textit{To summarize, \texttt{Raindrop} models with raw time series show comparable and oftentimes better performance than hand-crafted feature models across $23$ course iterations.}

\subsection{RQ2: Interpretability using TCAV}
In a second analysis, we investigated the use of concept activation vectors (TCAV) to create learner-centric interpretations of raw time series models.
%As described in the methodology section \textit{Concept-based Model Interpretation}, we aim to use TCAV to create learner-centric interpretations of raw time series models. Explainability for raw time series data is yet an unsolved problem; there has been minimal analysis of heatmaps of raw time series with more widely regarded explainability techniques (LIME, SHAP, Counterfactuals) used in the context of hand-crafted features \cite{swamy2022explainability}. 
We use the six dimensions of learning highlighted in Table \ref{tab:profiles} as concepts across our \texttt{Raindrop} model to interpret which aspects of the time series the model found important in determining student performance labels. In the following, we will examine these results on the DSP course at the $40\%$ predictive level to provide a comparative analysis to the interpretability study in \citet{swamy2022explainability}. This \texttt{Ripple} analysis can easily be extended to other courses or time series settings.

\begin{figure}[!t]
\centering
\includegraphics[width=0.9\linewidth, trim=1 1 1 35,clip]{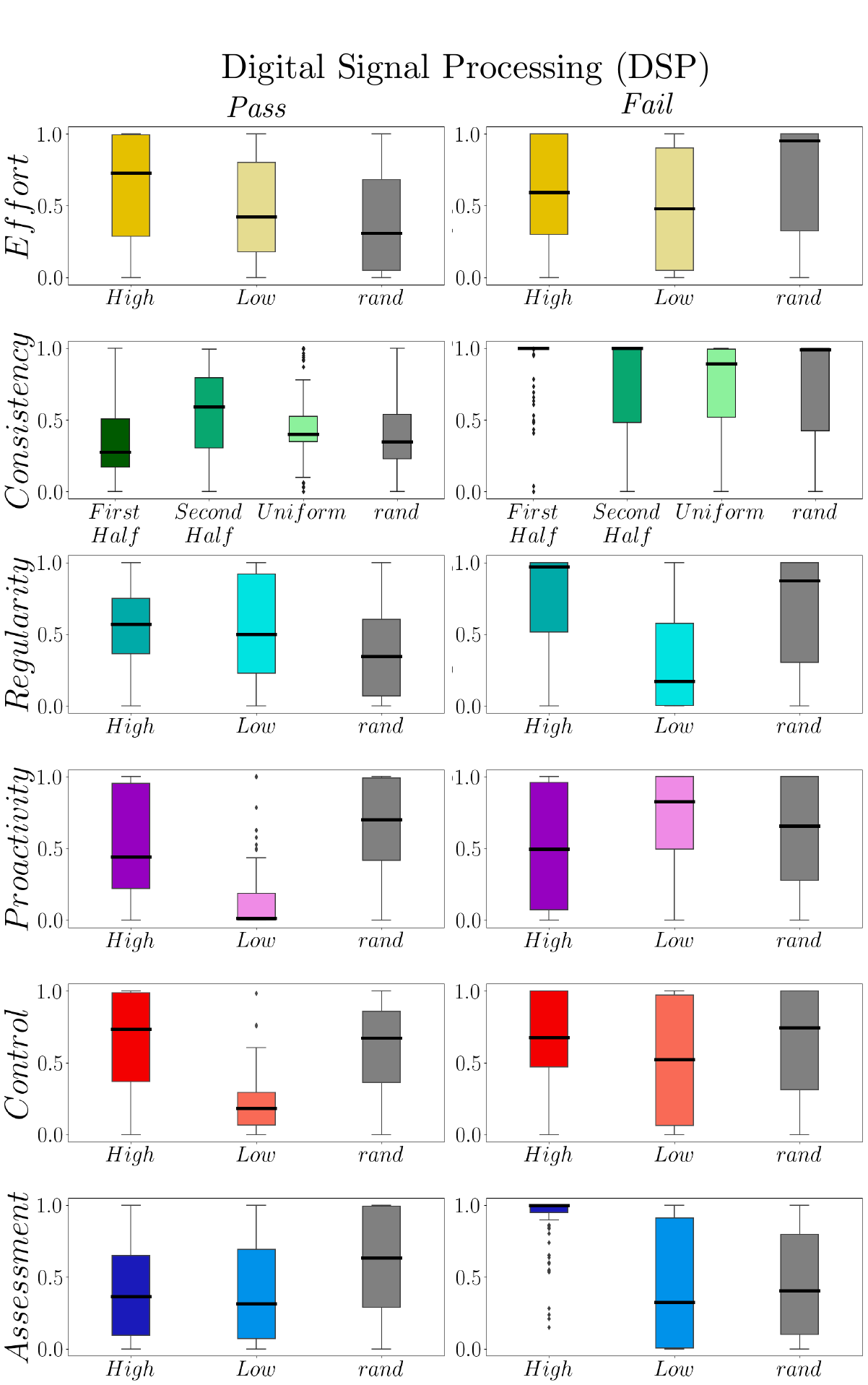}
\vspace{-1mm}
\caption{TCAV plots for early 40\% prediction on DSP to determine the importance of a concept to the model's predictions. For example, the last row shows a high TCAV score for high \textit{assessment} and a low TCAV score for random concepts for predicting student failure. This hints that the model is sensitive to \textit{assessment} for predicting  failure.}
\vspace{-3mm}
\label{fig:dsp40percent}
\end{figure}

Figure \ref{fig:dsp40percent} showcases TCAV concept sensitivity scores on DSP, across the model prediction classes of pass and fail. The significance of a specific pattern (i.e., \textit{uniform consistency}) can be analyzed in comparison to a random concept, defined by randomly choosing a subset of $100$ students without replacement $100$ times. TCAV scores are computed relative to other concepts in the same plot, so a low random concept score indicates that the other concepts are particularly important. We note that \textit{consistency} in the second half of the course for DSP is an important indicator of student success, more than consistent behavior in the first half or uniform consistency through the course. Interestingly, the model is sensitive to high \textit{assessment} scores when predicting student failure. Note that the score represents a sensitivity of the model only, with no indication on its direction. We hypothesize that performing well in assessments is actually a good indication for \textbf{not} failing, which is confirmed by the distribution of scores for \textit{assessment} measures when the model predicts failure. We further observe that the model is sensitive to high \textit{effort} when predicting passing.

% We extend this analysis to early success prediction using 40\% of the course Villes Africaines in Figure \ref{fig:va40percent}. \vini{add more analysis here} These analyses provide a global perspective of model performance.

\begin{figure}
\small
\centering
\includegraphics[width=0.95\linewidth, trim=1 1 1 1,clip]{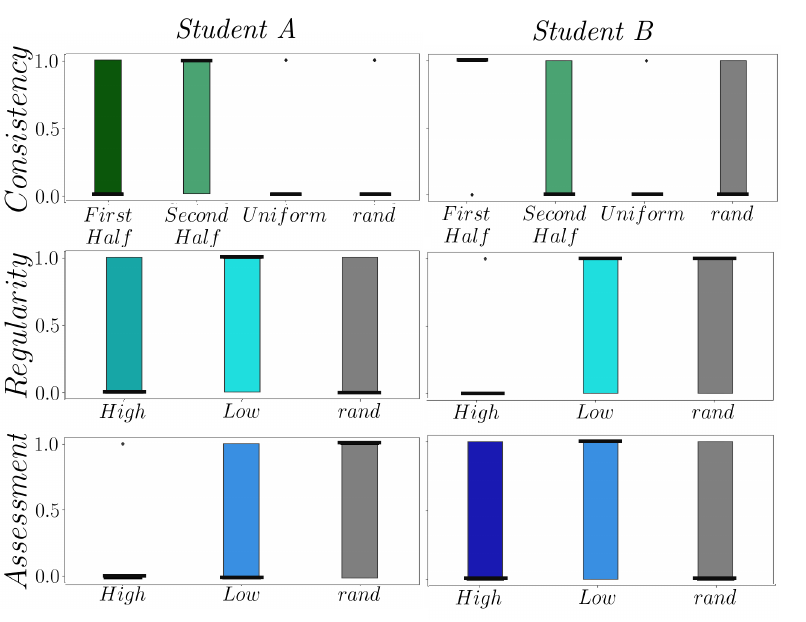}
\vspace{-2mm}
\caption{TCAV score plots for two students with differing behavioral characteristics in DSP. Student A is a high performer and student B is a low performer.}
\label{fig:top-bad}
\end{figure}

While this analysis provides a global perspective, it is also important to identify local, actionable insights based on early predictions. Our \texttt{Ripple} TCAV formulation enables the pipeline to provide interpretations for individual students. To observe local explanations, we examined the TCAV scores across a few interesting dimensions\footnote{Extended \texttt{Ripple} results can be found in our repository.} for a high performer (student A) and a low performer (student B). A student's performance was measured based on the average of all of the six dimensions from Table \ref{tab:profiles}, i.e. a highly performant student is among the top $t\%$ of all the dimensions combined and vice versa. The motivation for choosing these two case studies is two-fold: (i), we aimed to showcase a real-world use case for an educator to make individual student interventions, and (ii), we wanted to validate our interpretability methods on students who should have very different scores on TCAV dimensions because of their different levels of engagement. We computed TCAV plots across all behavioral dimensions for DSP on these two students and highlight three dimensions with interesting results in Figure \ref{fig:top-bad}. We saw that \textit{consistency} is indicative of performance for the high performer in the second half of the course and for the low performer in the first half of the course. We saw that low \textit{regularity} is a trait of the high performer while neither high nor low \textit{regularity} is important for student B (the random concept has a high TCAV score). Lastly, while \textit{assessment} is not important to student A, low \textit{assessment} score indicates the failing prediction of student B. 

\begin{figure}[!t]%
    \centering
    \subfloat[Regularity\label{fig:confidence-matrix-reg}]{
    \includegraphics[width=.92\linewidth, trim=0 4 4 4,clip]{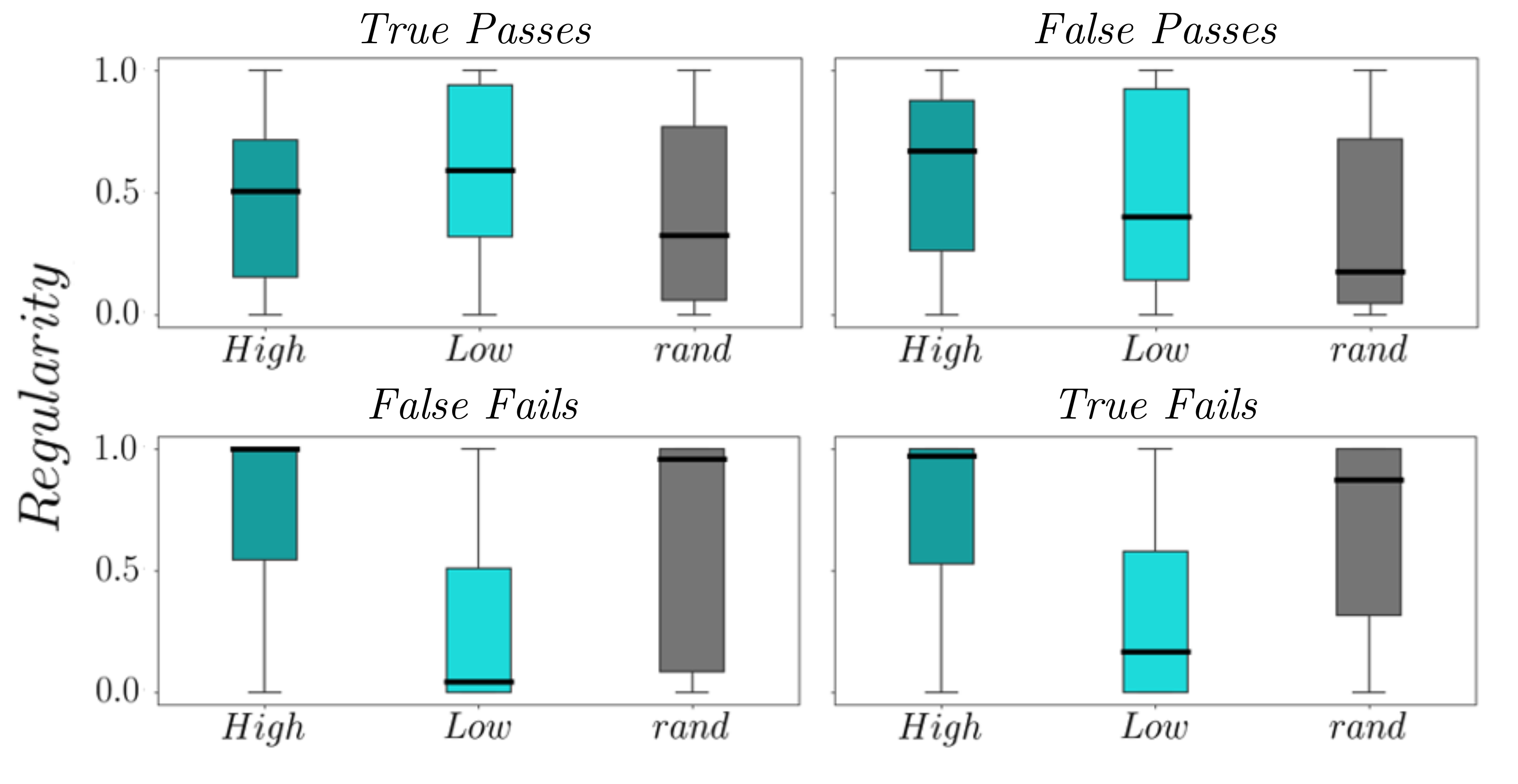}}
    \newline
    \subfloat[Effort\label{fig:confidence-matrix-eff}]{
    \includegraphics[width=.92\linewidth, trim=0 4 4 -2,clip]{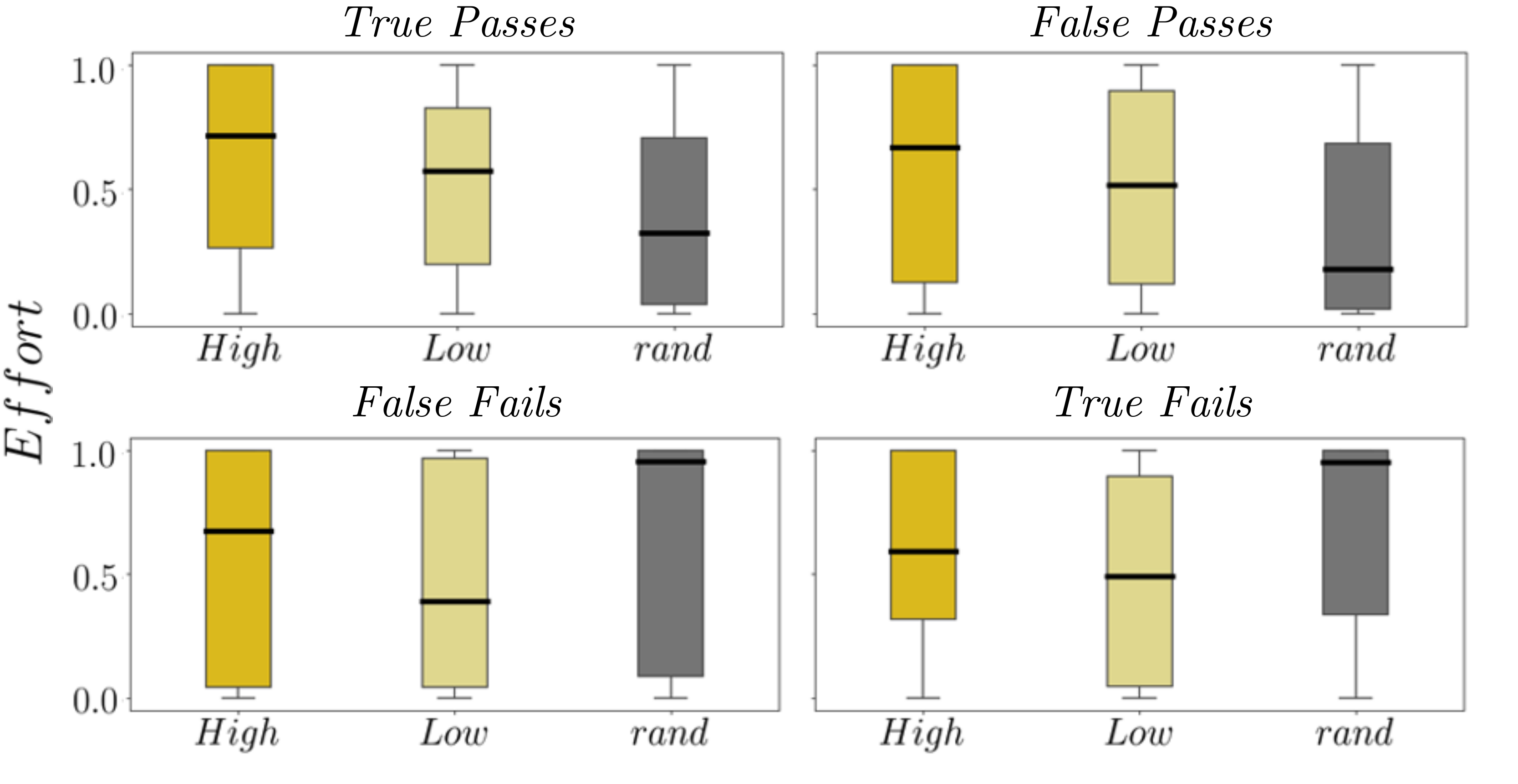}}
    \caption{Confusion matrix of TCAV plots across two dimensions for early 40\% prediction on DSP. \textit{True} or \textit{False} designate the model's correctness in prediction; \textit{True Passes} indicates the model predicted pass for students who did pass.}
    \vspace{-3mm}
    \label{fig:confidence-matrix}
\end{figure}

% 1. to showcase local predictions (real world use case for a teacher)
% 2. validate the interpretability methods on two differing students with high and low engagement

We are also interested in using TCAV to build intuition about how and why the model makes mistakes. Deep learning models have not seen the same uptick in adoption as traditional machine learning methods due to their lack of interpretability, which demonstrates practitioners' mistrust in model prediction. We aim to use TCAV to help educators trust time series prediction models by providing an avenue to understand model strengths and weaknesses.

In Figure \ref{fig:confidence-matrix}, we examined \textit{regularity} and \textit{effort} for DSP. Specifically, we investigated the four cases present in a traditional confusion matrix analysis, observing when the ground truth agrees with the model's predictions (true positives and true negatives) and when they disagree (false positives, false negatives). For \textit{regularity} (Figure \ref{fig:confidence-matrix-reg}), we observed an inverse relationship between true positives and false positives. When the model made a correct prediction for passing students, it was more sensitive to low \textit{regularity} and when the model predicted a false positive, it was more sensitive to high \textit{regularity}. We can hypothesize that when \texttt{Ripple} incorrectly identified a student as passing, this student has high \textit{regularity} scores and that tricked the model into getting it wrong. For false and true negatives, we see high random concept values for \textit{regularity}. We can infer that high and low \textit{regularity} are not important concepts for predicting failure. Similarly, in Figure \ref{fig:confidence-matrix-eff} for the \textit{effort} dimension, we can infer that neither high nor low \textit{effort} concepts contributed to predicting failure. However, for the passing case, we see a parallel relationship across true and false positives:  the model likely always predicted that students with high \textit{effort} pass the course (and sometimes this was incorrect). Going further to examine the distributions of the dimension values for each of these student subsets would enable educators to validate these hypotheses. Through confusion matrix plots, it is possible to examine different subsets of the student population  to analyze model strengths and weaknesses in detail.

\textit{Overall, \texttt{Ripple} enables globally interpretable feedback on the scale of thousands of students in a course and locally actionable feedback on the scale of a specific student without requiring a model built on hand-crafted features.}

%%%%%%%%%Conclusion%%%%%%%%%%%%%%%%%%%%%%%%%%%%%%%%%%%%%%%%%%%%%%%%%%%%%
\section{Conclusion}
In this paper, we introduced \texttt{Ripple}, a pipeline to make predictions from raw time series and interpret them with learner-centric concepts. We demonstrated that the performance of the underlying \texttt{Raindrop} models is comparable and often considerably better than hand-crafted feature models. Furthermore, we showed that it is feasible to define human-friendly concepts and make intuitive and actionable interpretations of model behavior. We also suggested the use of TCAV interpretability analysis to build trust in models.

The novelty of this work lies in combining the state-of-the-art AI advances in time series modeling and interpretability together and examining their implications for educational data. Educators can now get granular insights about their course (global scale) and individual students (local scale) that are based on concepts that they specify as important. The flexibility of interpretation that TCAV offers (user-specified concepts, granularity of insights, accuracy in directly using model activations) applies to any educational time series prediction setting and beyond, making it ideal for any scenario where there is direct impact on humans.
% add a few sentences about implications here

% limitations
In future work, we plan to run experiments on a larger dataset, with a more international audience and more interaction modalities (i.e., flipped classrooms, simulation data). %extracting TCAV for a course to receive interpretable insights does require some effort (although we mostly used simple statistics). 
We also hope to extend \texttt{Ripple} using transfer learning across courses and to provide generalized concept vectors to be used for a multi-course model. This would allow to use raw time series input with deep learning models and maintain both accuracy and interpretability without extra effort. Finally, \texttt{Raindrop} can be further optimized for efficiency.

\bibliography{aaai23}
\end{document}